# Semi-supervised Hashing for Semi-Paired Cross-View Retrieval

Jun Yu[1], Xiao-Jun Wu[1*], Josef Kittler[2]

[1] School of IoT Engineering, Jiangnan University, 214122, Wuxi, China
[2] Center for Vision, Speech and Signal Processing(CVSSP), University of Surry, GU2 7XH Guildford, UK
yujunjason@aliyun.com, xiaojun_wu_jnu@163.com, j.kittler@surrey.ac.uk

*Abstract*—Recently, hashing techniques have gained importance in large-scale retrieval tasks because of their retrieval speed. Most of the existing cross-view frameworks assume that data are well paired. However, the fully-paired multiview situation is not universal in real applications. The aim of the method proposed in this paper is to learn the hashing function for semi-paired cross-view retrieval tasks. To utilize the label information of partial data, we propose a semi-supervised hashing learning framework which jointly performs feature extraction and classifier learning. The experimental results on two datasets show that our method outperforms several state-of-the-art methods in terms of retrieval accuracy.

## I. INTRODUCTION

The explosive growth of multimedia data creates more challenges in information retrieval. Hashing, which is an effective feature representation of data, has received increasing attention in multimedia data analysis, computer vision and related areas due to its low storage and high speed of processing [1].

Among many hashing techniques, locality Sensitive Hashing (LSH) [2] is a popular data-independent method of generating hash code by means of random projections. It inspired the popular Kernelized Locality Sensitive Hashing (KLSH) [3] and Shift-invariant Kernels hashing (SIKH) [4] methods. But, these algorithms need long hash code to achieve high performance. This increases the memory consumption and storage cost. To tackle the problem, many data-dependent methods in which hash functions are learned from a given training dataset were proposed for single view, such as Discrete Graph Hashing (DGH) [5], Scalable Graph Hashing (SGH)[6], Anchor Graph-based hashing (AGH) [7], Iterative Quantization (ITQ) [8], Semi-supervised hashing (SSH) [9], Latent factor models for supervised hashing (LFH)[10], Supervised discrete hashing (SDH)[11], etc. These data-dependent methods achieve comparable or even better accuracy with shorter binary codes, compared with data-independent methods. However, these achievements do not directly apply to multi-view situation.

In many real application systems, some objects can be represented by two or more kinds of features. For example, a webpage in Internet can be represented by text, image, video, and hyper-link. The kind of data similar to webpage is reffered to as multi-view data. In general, multi-view Hashing is an important requirement in many practical applications. There are two major categories of existing multi-view hashing methods. i.e. Multiview Hashing and Cross-view Hashing. By leveraging auxiliary views, Multiview Hashing, such as Multiple feature hashing (MFH) [12], Composite hashing (CH) [13], etc, promises to learn better codes than single-view. But, these methods must satisfy the condition that all data views are available. Unlike Multiview Hashing, Cross-view Hashing is designed to conduct cross-view retrieval. A query from one view can retrieve the relevant results from another view. Cross-view Hashing can also be divided into Supervised Cross-view Hashing and Unsupervised Cross-view Hashing. Most of the Unsupervised Cross-view methods depend on canonical correlation analysis (CCA). Specifically, they tranform multiple feature views into a common latent subspace in which the correlation among all views is maximized. Besides, Semi-paired hashing (SPH) [14] not only keeps the correlation between two views but also preserves structural similarity within a view. Semi-paired Discrete Hashing (SPDH) [15] implements the mapping by preserving the intrinsic similarities of semi-paired data based on the idea of anchor pair. Supervised Cross-view Hashing which can use the semantic label for hashing learning has achieved promising results. An example is Semantic Correlation Maximization (SCM) [16] which integrates semantic labels into hashing learning.

Supervised Cross-view hashing often needs a lot of label information to train a robust hashing function. However, it is time-consuming and labor-intensive to collect labeled samples. On the contrary, massive unlabeled data is obtained easily in many applications. In order to leverage substantial unlabeled samples and limited labeled samples,and take the universality of semi-paired data in real-world into account, we propose a Semi-supervised hashing learning framework for the semi-paired cross-view retrieval problem. The learning process of our model consists of two stages: Firstly, the mapping from the original space to a commom subspace is learned by means of relaxing an optimization problem. Secondly, we minimize the quantization error to refine the hashing function. Compared with [17], our model is unique in terms of the objective function design for a multi-class problem in the multi-views scenario. The main contributions of our framework are given as follows:

*(1) In multi-view scenarios, our method integrates feature extraction and classifier learning into a joint framework to*

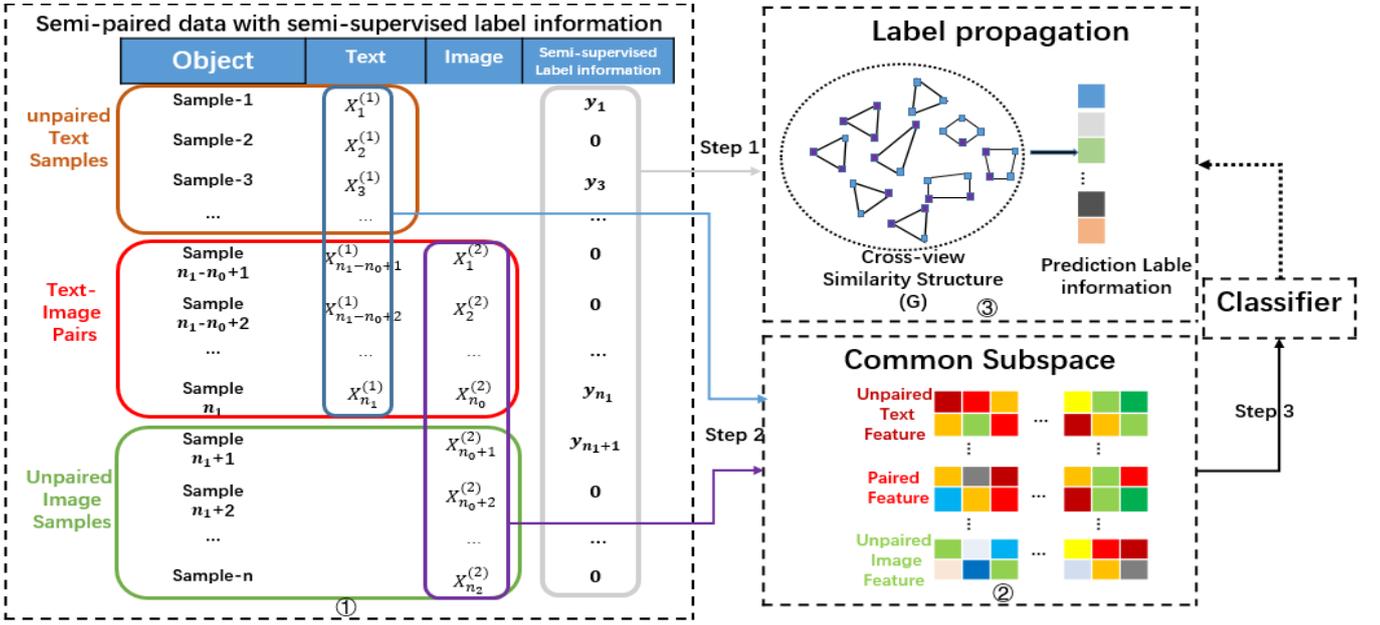

Fig. 1. Illustration of our framework (better viewed in color). Training dataset consists of *n* samples which are divided into three parts: (i) Unpaired Text Samples framed by orange in ① are represented only by text feature; (ii) Text-Image Pairs framed by red in ① are represented by both text feature and Image feature; (iii) Unpaired Image Samples framed by green in ① are represented only by image feature. Step 1: Prediction label should be consistent with both the ground true label of labeled data (framed by gray in ①) and the whole graph G over all data for label propagation. Step2:Text feature data framed by blue in ① and Image feature data framed by purple in ① are projected a common subspace respectively. Step 3: data points in common latent subspace are classified.

learn the hashing function mapping the original data to the Hamming space.

(2) With the help of anchor graph, we construct cross-view graph for label propagation.

The paper includes four sections. Related research status is described in Section I. In Section II, we formulate the research problem and develop our model and optimization process. The experimental results and analysis are depicted in Section III. The conclusions are presented in Section IV.

## II. METHODOLOGY

### A. Problem statement

For simplicity, we formulate the problem with two views, i.e. Image-view and Text-view. As shown in Figure.1, let $X^{(1)}$ denote Text-view and $X^{(2)}$ signify Image-view. Specifically, $X^{(1)} = [x_1^{(1)}, \ldots, x_{n_1-n_0}^{(1)}, x_{n_1-n_0+1}^{(1)} \ldots, x_{n_1}^{(1)}]^T$,

$X^{(2)} = [x_1^{(2)}, \ldots, x_{n_0}^{(2)}, x_{n_0+1}^{(2)} \ldots, x_{n_2}^{(2)}]^T$, where $x^{(1)} \in R^{d_1}$, $x^{(2)} \in R^{d_2}$ ($d_1 \neq d_2$), $n_i$ and $d_i$ ($i=1,2$) are the number of samples and dimensionality in the $i^{th}$ view respectively. Here we consider $\{X_{n_1-n_0+i}^{(1)}, X_i^{(2)}\}_{i=1}^{n_0}$ are paired data points, that is, the last $n_0$ samples in the text view and the first $n_0$ samples in the image view come from the same $n_0$ objects, where $n_0$ is the total number of pairs. Suppose that *m* samples selected randomly from the training set are labeled and the rest are unlabeled data. We define a label matrix $Y=[y_1, \ldots y_m, y_{m+1}, \ldots y_n]^T$, where $y_i|_{i=1}^n \in \{0,1\}^c$ and *c* is the total number of classes; $y_{ij}$ is the $j^{th}$ element of $y_i$, $y_{ij} = 1$ if the $i^{th}$ sample belongs to the $j^{th}$ class and $y_{ij} = 0$ otherwise. We assume $y_i$ is zero vetor if the label information is not available. Samples are zero-centered for each view feature, i.e. $\sum_{i=1}^{n_1} X_i^{(1)} = 0$, $\sum_{i=1}^{n_2} X_i^{(2)} = 0$. The aim of the paper is to obtain a mapping function from the original data space to Hamming space for two different views, that is, $f: R^{d_1} \to \{1,0\}^r$ and $g: R^{d_2} \to \{1,0\}^r$ where *r* is the length of hash code. Here, the hash function is set into the following form:

$$f(X^{(1)}) = sign(X^{(1)}Q^{(1)})$$

$$g(X^{(2)}) = sign(X^{(2)}Q^{(2)}) \quad (1)$$

where sign(.) signifies sign function, and $Q^{(i)} \in R^{d_i \times r}$ ($i = 1,2$) are the projection matrices for the $i^{th}$ view.

### B. Model

The basic idea of our method is to try to learn the hash function mapping different views into a common Hanming space. The objective function is formulated as a multi-class classification problem. i.e.

$$\min_{W,Q^{(i)}} \sum_{i=1}^2 \theta^{(i)} ||Y^{(i)} - sgn(X^{(i)}Q^{(i)})W||_F^2 + \beta||W||_F^2 \quad (2)$$

where $\theta^{(i)}(i = 1,2)$ is a variable that indicates the relative importance of the $i^{th}$ view in the learning process and $Y^{(i)} \in R^{n_i \times c}$ is the semi-supervised label matrix for the $i^{th}$ view. Owing to the data being semi-supervised data, we introduce a label prediction matrix F=$[F_1, \ldots, F_n]^T$, where $F_i \in R^c$, to make the best use of available label information. F should be consistent with the ground truth label for the labeled data and graph G over all data for label propagation [18][19]. This can be obtained by optimizing the cost function in (3).

$$\min_F \sum_{p=1}^c \left[ \frac{1}{2} \sum_{i,j=1}^n (F_{ip} - F_{jp})^2 S_{ij} + \sum_{i=1}^n U_{ii}(F_{ip} - Y_{ip})^2 \right] \quad (3)$$

where $F_{ip}$ denotes the $p^{th}$ element of $F_i$ ($i=1,...n$). $U_{ii} = \infty$ if the label of the $i^{th}$ sample is available and $U_{ii} = 0$ otherwise, $S_{ij}$ is the $i^{th}$ row and the $j^{th}$ colum of similarity matrix $S \in R^{n \times n}$ and denotes the similarity between sample $i$ and sample $j$. It is hard to directly compute the similarity among data points represented by different views. In this paper, we adopt an effective anchor graph construction approach [20] to analyze the similarity among semi-paired data. We randomly select $m_0$ pairs potential as anchor pairs from the paired samples. The similarity matrix $Z \in R^{n \times m}$ between samples and anchors can be defined as:

$$Z_{ij} = \begin{cases} \frac{\exp(\Phi(x_i, \mu_j))}{\sum_{j \in [i]} \exp(\Phi(x_i, \mu_j))}, & \forall j \in [i] \\ 0 & otherwise \end{cases} \quad (4)$$

where $[i]$ are the k-nearest anchors of the $i^{th}$ sample $x_i$, $\Phi(x_i, \mu_j) = -||x_i - \mu_j||^2 / \sigma^2$, and $\mu_j$ is an anchor which is from the set of k-nearest anchors. To normalize each row, $\Lambda = \text{diag}(Z^T 1) \in R^{m \times m}$ is used in (5). Then, the similarity matrix S can be calculated as

$$S = Z\Lambda^{-1} Z^T \quad (5)$$

As seen in [17], the function in (3) can be converted into

$$\min_F Tr(F^T L F) + Tr[(F - Y)^T U(F - Y)] \quad (6)$$

where $Tr(.)$ denotes the trace of matrix, $L = I - S$ is the Laplacian matrix. As paired view data from the same object should be close to each other in the Hamming space, we acquire

$$\min_{f,g} \sum_{j=1}^{n_0} ||f(X^{(1)}_{n_1 - n_0 + j} Q^{(1)}) - g(X^{(2)}_j Q^{(2)})||^2 \quad (7)$$

Integrating (2), (6) and (7) into a joint framework, we can optimize classifier learning and feature representation simultaneously. The joint optimization problem is given as follows:

$$\min_{F,Q^{(i)},W,\theta^{(i)}} Tr(F^T L F) + Tr[(F - Y)^T U(F - Y)]$$
$$+ \sum_{i=1}^2 \theta^{(i)} ||F^{(i)} - sgn(X^{(i)} Q^{(i)})W||_F^2 + \beta ||W||_F^2$$
$$+ \gamma \sum_{j=1}^{n_0} ||sgn(X^{(1)}_{n_1 - n_0 + j} Q^{(1)}) - sgn(X^{(2)}_j Q^{(2)})||^2$$
$$\text{s.t.} \sum_{i=1}^2 \theta^{(i)} = 1, \theta^{(i)} > 0, i = 1,2 \quad (8)$$

where $\gamma$ and $\beta$ are non-negative adjustable parameters.

*C. Relaxation and optimization*

It is difficult directly to optimize the problem with a binary function in (8). We relax the objective function to make it tractable computationally. We define four view-specific element selection matrices $T^{(1)} = [1_{n_1 \times n_1}, 0_{n_1 \times (n - n_1)}]$, $T^{(2)} = [0_{n_2 \times (n_1 - n_0)}, 1_{n_2 \times n_2}]$, $M^{(1)} = [0_{n_0 \times (n_1 - n_0)}, 1_{n_0 \times n_0}]$, $M^{(2)} = [1_{n_0 \times n_0}, 0_{n_0 \times (n_2 - n_0)}]$. The relaxed objective function can be formulated as:

$$\min_{F,Q^{(i)},W,\theta^{(i)}} Tr(F^T L F) + Tr[(F - Y)^T U(F - Y)]$$
$$+ \sum_i^2 \theta^{(i)} ||T^{(i)} F - X^{(i)} Q^{(i)} W||_F^2 + \beta ||W||_F^2$$
$$+ \gamma ||M^{(1)} X^{(1)} Q^{(1)} - M^{(2)} X^{(2)} Q^{(2)}||_F^2$$
$$\text{s.t.} \sum_{i=1}^2 \theta^{(i)} = 1, \theta^{(i)} > 0, i = 1,2 \quad (9)$$

We use ADMM algorithm to solve the relaxed problem in (9) through alternating optimization.

*(1) Optimization of F: keeping terms relating to F*

$$\min_F Tr(F^T L F) + Tr[(F - Y)^T U(F - Y)]$$
$$+ \sum_i^2 \theta^{(i)} ||T^{(i)} F - X^{(i)} Q^{(i)} W||_F^2 \quad (10)$$

Setting the derivative of (10) with respect to $F$ to zero, we obtain the optimal $F$ as

$$(L + U)F - UY + \sum_{i=1}^2 \theta^{(i)} [T^{(i)^T} T^{(i)} F - T^{(i)^T} X^{(i)} Q^{(i)} W] = 0 \quad (11)$$

$$\Rightarrow F = (L + U + \sum_{i=1}^2 \theta^{(i)} T^{(i)^T} T^{(i)})^+ [\sum_{i=1}^2 \theta^{(i)} T^{(i)^T} X^{(i)} Q^{(i)} W + UY] \quad (12)$$

where the superscript + denotes pseudoinverse of a matrix.

*(2) Optimization of W: keeping terms relating to W*

$$\min_W \sum_{i=1}^2 \theta^{(i)} ||T^{(i)} F - X^{(i)} Q^{(i)} W||_F^2 + \beta ||W||_F^2 \quad (13)$$

Letting the derivative of (13) with respect to W equal to zero, we can obtain the following closed form solution as (15):

$$\sum_{i=1}^2 \theta^{(i)} [Q^{(i)^T} X^{(i)^T} X^{(i)} Q^{(i)} W - Q^{(i)^T} X^{(i)^T} T^{(i)} F]$$
$$+ \beta W = 0 \quad (14)$$

$$\Rightarrow W = \left( \sum_{i=1}^2 \theta^{(i)} Q^{(i)^T} X^{(i)^T} X^{(i)} Q^{(i)} + \beta I \right)^{-1}$$
$$\left( \sum_{i=1}^2 \theta^{(i)} Q^{(i)^T} X^{(i)^T} T^{(i)} F \right) \quad (15)$$

*(3) Optimization of $Q^{(i)}$: keeping terms relating to $Q^{(i)}$*

$$\min_{Q^{(i)}} \sum_{i=1}^2 \theta^{(i)} ||T^{(i)} F - X^{(i)} Q^{(i)} W||_F^2$$
$$+ \gamma ||M^{(1)} X^{(1)} Q^{(1)} - M^{(2)} X^{(2)} Q^{(2)}||_F^2 \quad (16)$$

Taking the derivate of (16) with respect to $Q^{(1)}$ and $Q^{(2)}$ equal to zero, results in

$$\gamma(X^{(1)^T}X^{(1)})^+ X^{(1)^T} M^{(1)^T} M^{(1)} X^{(1)} Q^{(1)} + Q^{(1)} \theta^{(1)} WW^T = \\ (X^{(1)^T}X^{(1)})^+ (\theta^{(1)} X^{(1)^T} T^{(1)} FW^T + \\ \gamma X^{(1)^T} M^{(1)^T} M^{(2)} X^{(2)} Q^{(2)}) \quad (17)$$

$$\gamma(X^{(2)^T}X^{(2)})^+ X^{(2)^T} M^{(2)^T} M^{(2)} X^{(2)} Q^{(2)} + Q^{(2)} \theta^{(2)} WW^T = \\ (X^{(2)^T}X^{(2)})^+ (\theta^{(2)} X^{(2)^T} T^{(2)} FW^T + \\ \gamma X^{(2)^T} M^{(2)^T} M^{(1)} X^{(1)} Q^{(1)}) \quad (18)$$

which is consistent with the Sylvester equation.

*(4) Optimization of $\theta^{(i)}$: keeping terms relating to $\theta^{(i)}$*

$$\min_{\theta^{(i)}} \sum_{i=1}^{2} \theta^{(i)} \pi^{(i)} + \lambda ||\Theta||_2^2$$
$$s.t. \sum_{i=1}^{2} \theta^{(i)} = 1, \theta^{(i)} > 0, i = 1,2 \quad (19)$$

where $\pi^{(i)} = ||T^{(i)}F - X^{(i)}Q^{(i)}W||_F^2$, $\Theta = [\theta^{(1)}, \theta^{(2)}]^T$. The second part of Eq.19 is a regularization term to exploit the complementary information of the two views. Equation (19) is a quadratic optimization problem which can be solved easily by any existing algorithm. $\lambda$ is non-negative coefficient for controlling the smoothness of $\Theta$.

The iterative method to solve the relaxed objective function in (9) is described in Algorithm 1. The iteration process is repeated until the algorithm converges. After acquiring the projection matrix $Q^{(i)}$, we can quantize the projected data into binary codes. To improve the hashing performance, we introduce the ITQ algorithm [8] to reduce quantization error. For different two views, quantization is achieved in the same way by optimizing E.q. 20.

$$\min_{B^{(i)}, R^{(i)}} ||B^{(i)} - X^{(i)} Q^{(i)} R^{(i)}||_F^2 \quad (20)$$
$$s.t \; B^{(i)} \in \{1,0\}^{n_i \times r}, R^T R = I$$

where $B^{(i)}$ denotes the hash code matrix of the $i^{th}$ view, $R^{(i)} \in R^{r \times r}$ is a transformation matrix with the property of orthogonality, employed to align with the hypercube $\{0,1\}^{n_i \times r}$ by means of rotating the data. There are two steps to optimize E.q.20 by the alternating algorithm:

Step 1: Fix $R^{(i)}$ and update $B^{(i)}$. The objective function generates a closed form solution as :

$$B^{(i)} = sgn(X^{(i)} Q^{(i)} R^{(i)}) \quad (21)$$

Step 2: Fixed $B^{(i)}$ and update $R^{(i)}$. The solution can be obtained efficiently with the singular value decomposition.

$$R^{(i)} = V^{(i)} U^{(i)^T} \quad (22)$$

The framework proposed in the paper consist of a two-stage mechanism: (1) Projected matrices $Q^{(i)}$ are learned from the relaxed objective function via algorithm 1; (2) The orthogonal transformation matrix R is obtained through minimizing the quantization error in E.q.20. Thus, the hash function $H^{(i)}$ of the $i$th view ($i$=1,2) can be defined as:

$$H^{(i)}(x^{(i)}) = sgn(x^{(i)} Q^{(i)} R^{(i)}) \quad (23)$$

where $x^{(i)} \in R^{1 \times d_i}$ is a randomly selected sample in the $i$th view.

---

**Algorithm 1:** Algorithm for solving the relaxed objective function in (9)

**Input**: trainning set $X^{(i)} \in R^{n_i \times d_i}(i = 1,2)$, semi-supervised label matrix $Y \in \{0,1\}^{n \times c}$ and parameters $\beta, \gamma$
**Output**: prediction matrix $F \in R^{n \times c}$, weighted matrix $W \in R^{r \times c}$ and projection matrix $Q \in R^{d_i \times r}$.
1: Initialize F, W, $Q^{(i)}, \theta^{(i)}$
2: Calculate S according to (5)
3: Calculate Laplacian matrix L and diagonal matrix U.
4: **Repeat**
5:     Update F according to (12);
6:     Update W according to (15);
7:     Update $Q^{(1)}$ according to (17);
8:     Update $Q^{(2)}$ according to (18);
9:     Update $\theta^{(i)}$ by solving (19);
10: **Until** convergence
11: Return F, W, $Q^{(1)}, Q^{(2)}$.

---

## III. EXPERIMENTAL RESULTS

In this section, we present extensive experiments to evaluate the proposed method. We compare different methods on two publicly available datasets: MIRFlickr, and Wiki.

### A. Datasets

MIRFlickr consists of 25000 original images collected from Flickr website. Each image is classified into 24 classes. We randomly select 7000 samples whose textual tags are not empty for our experiments. The image view is quantized into 150-dimensional edge histogram feature vector and the text view is represented as 500-dimensional vector derived from the bag-of–words vector by adopting PCA. 70% of the data points are used as the training set and, the rest of the database as query set.

Wiki contains 2866 multimedia documents crawled from Wikipedia. A 128-dimensional SIFT histogram vector and a 10-dimensional feature vector generated by latent Dirichlet allocation are used to represent each image and text respectively. We take 75% of the dataset to form the training set and the remaining 25% as a test set.

### B. Experimental Setting

The performance of the proposed method is evaluated on two kinds of retrieval task, i.e. image to texts and text to image. They are represented by I → T and T → I respectively. The number of anchors $m_0$ is equal to 10% of pairs if the number exceeds fifty, fifty otherwsie. The value of k for the nearest anchor is directly set to $m_0$. Samples with label cover all categories in our experiments. Our method is compared with most the popular methods including SCM-orth, CCA, and SPDH. Since the soure code of SPDH is not publicly available, we implemented it ourselves. CCA and SCM-orth are kindly provided by their authors. Note that SPDH and CCA are unsupervised methods and SCM-orth is a supervised method for semi-paired cross view retrieval. We use the Mean Average Precision (MAP) as the indicator of the retrieval performance. The Average Precision for a query $q$ is formulated as E.q.24

$$AP(q) = \frac{1}{l_q}\sum_{m=1}^{R} P_q(m)\delta_q(m) \qquad (24)$$

where $l_q$ denotes the correct statistics of top $R$ retrieval results; $P_q(m)$ is the accuracy of top $m$ retrieval results; and if the result of position $m$ is right, $\delta_q(m)$ is equal to one or zero otherwise. We set R=50 in our experiment.

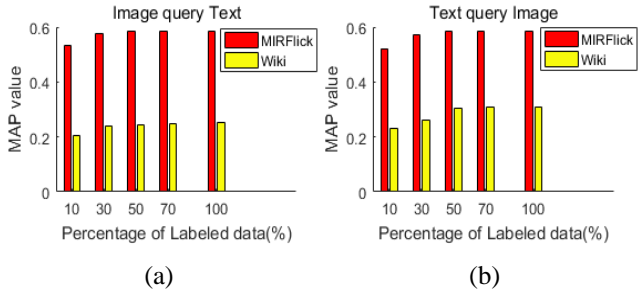

Fig. 2. The performance variation of our method with respect to the percentage of labeled data in the training set for two datasets when the length of hash code and the precentage of pairs are fixed to 32 and 50% respectively. (a) I → T, (b) T → I.

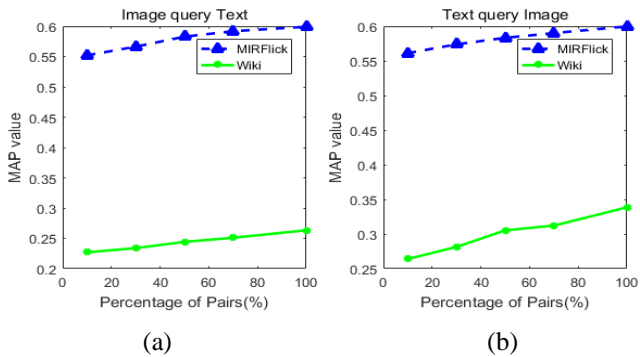

Fig. 3. The performance variation of our method with respect to the percentage of pairs in the training data obtained on two datasets when the length of hash code and the precentage of labeled data are 32 and 50% fixed respectively. (a) I → T, (b) T → I.

We set the possible values of $\beta$ and $\gamma$ in the range of {0.01, 0.1, 1, 10, 100,1000} empirically, and the best results are recorded in this paper. The MAP performance variation of our method with respect to $\beta$, $\gamma$ in Fig.5, when using 64-bit hash code on wiki and MIRFlickr. we can see that choosing $\beta$, $\gamma$ have influence on the Map performance.

One advantage of our method is that it fully utilizes the semi-supervised semantic information to process multi-view data in the semi-paired scenario. The convergence curves of the algorithm on the two datasets are plotted in Fig.4, which show the aglorithm can converge in about 15 iterations. SPDH is also a state-of-art method to deal with semi-paired samples. But it is an unsupervised method which exploits a common latent space by constructing a similarity graph between views according to

## C. Performance Evaluation

TABLE 1. Comparative map results on MIRFlickr

| Task | Method | Code Length | | |
|---|---|---|---|---|
| | | 16 | 32 | 64 |
| Image query Text (I → T) | Ours(100%) | **0.5897** | **0.5991** | **0.6079** |
| | Ours(50%) | 0.5798 | 0.5828 | 0.5871 |
| | SPDH(100%) | 0.5698 | 0.5719 | 0.5745 |
| | SPDH(50%) | 0.5422 | 0.5499 | 0.5486 |
| | SCM-orth | 0.5552 | 0.5570 | 0.5546 |
| | CCA | 0.5559 | 0.5583 | 0.5590 |
| Text query Image (T → I) | Ours(100%) | **0.5881** | **0.5997** | **0.6047** |
| | Ours(50%) | 0.5638 | 0.5833 | 0.5962 |
| | SPDH(100%) | 0.5706 | 0.5736 | 0.5762 |
| | SPDH(50%) | 0.5382 | 0.5413 | 0.5486 |
| | SCM-orth | 0.5460 | 0.5473 | 0.5521 |
| | CCA | 0.5377 | 0.5456 | 0.5476 |

TABLE 2. Comparative map results on Wiki

| Task | Method | Code Length | | |
|---|---|---|---|---|
| | | 16 | 32 | 64 |
| Image query Text (I → T) | Ours(100%) | **0.2429** | **0.2633** | **0.2529** |
| | Ours(50%) | 0.2219 | 0.2442 | 0.2328 |
| | SPDH(100%) | 0.2179 | 0.2264 | 0.2319 |
| | SPDH(50%) | 0.2077 | 0.2127 | 0.2246 |
| | SCM-orth | 0.2023 | 0.2289 | 0.2068 |
| | CCA | 0.2136 | 0.2583 | 0.2396 |
| Text query Image (T → I) | Ours(100%) | 0.2979 | **0.3306** | 0.3186 |
| | Ours(50%) | 0.2663 | 0.3053 | 0.2963 |
| | SPDH(100%) | 0.2701 | 0.3019 | **0.3294** |
| | SPDH(50%) | 0.2501 | 0.2961 | 0.3037 |
| | SCM-orth | 0.2973 | 0.2641 | 0.2157 |
| | CCA | **0.3228** | 0.2610 | 0.2230 |

the idea of anchor pairs. We compare our method with SPDH under two different pair settings, i.e., 100% and 50%. As shown in Table 1 and Table 2, the map results of our methods are better than SPDH on MIRFlickr and Wiki by approximately 2% when the percentage of labeled data is fixed at 50%. CCA is also an unsupervised method in which the correlation among views is maximized. Moreover, SCM-orth is a supervised method which constructs the semantic similarity using the label vectors. However, the two methods are designed to deal with the data paired in their entirety. In Table 1 and Table 2, we can find that our method outperforms CCA and SCM-orth even when the percentage of pairs is 100%.

To exploit the impact of the number of labeled data on accuracy, we adjust the proportion of labeled data when we fixed the the length of hash code and the precentage of pairs is either 32 or 50% respectively. Figure 2 shows the map value is improved only about 1% when the propotion of labeled data range from 50% to 100%. This indicates that our method is more suitable for real application because it is is difficult to collect labeled data.

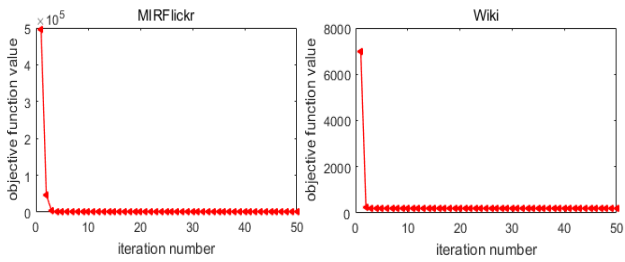

Fig. 4. Convergence curves of Algorithm 1 on MIRFlickr and Wiki.

Our method can be used in semi-paired scenarios. The map results have been obtained when we change the percentage of paired data from 10% to 100% in Figure 3. The performance degrades only slowly when the percentage of paired data reduces.

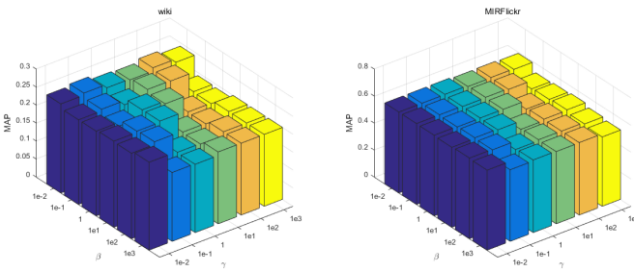

Fig. 5. The influence for results with respect to α and β using 64-bit hash code onWiki and MIRFlickr.

## IV. CONCLUSIONS

In this paper, we presensted a novel framework to learn hashing for cross-view data. The new algorithm jointly optimizes feature representation and the classifier parameters to learn high quality binary code. In the learning process, the framework exploits any label information provided by the training data set, even if partial to better adapt to real application. The experimental results demonstrate that our method is superior to other several popular methods in accuracy. In the future, we plan to add into our framework structure preservation and manifold embedding in the projection stage of our method.

## V. ACKNOWLEDGMENTS

THE PAPER IS SUPPORTED BY THE NATIONAL NATURAL SCIENCE FOUNDATION OF CHINA(GRANT NO.61373055, 61672265, 61603159), UK EPSRC GRANT EP/N007743/1, MURI/EPSRC/DSTL GRANT EP/R018456/1, AND THE 111 PROJECT OF MINISTRY OF EDUCATION OF CHINA (GRANT NO. B12018).